\documentclass{ieeeaccess}
\usepackage{cite}
\usepackage{amsmath,amssymb,amsfonts}
\usepackage{algorithmic}
\usepackage{diagbox}
\usepackage{array}
\usepackage{stfloats}
\usepackage{graphicx}
\usepackage{textcomp}
\usepackage{float}
\def\BibTeX{{\rm B\kern-.05em{\sc i\kern-.025em b}\kern-.08em
    T\kern-.1667em\lower.7ex\hbox{E}\kern-.125emX}}
\begin{document}
\history{Date of publication xxxx 00, 0000, date of current version xxxx 00, 0000.}
\doi{10.1109/ACCESS.2023.0322000}

\title{Design and control of a robotic payload stabilization mechanism for rocket flights}
\author{Utkarsh Anand\authorrefmark{1}\authorrefmark{*} | Diya Parekh\authorrefmark{2}\authorrefmark{*} | Thakur Pranav G. Singh\authorrefmark{3}\authorrefmark{*} | Hrishikesh S. Yadav\authorrefmark{4} | Ramya S. Moorthy\authorrefmark{5} | Srinivas G.\authorrefmark{6}}

\address[1]{Department of Electrical \& Electronics Engg., Manipal Institute of Technology, Karnataka, India - 576104 (email: utkarshanand221@gmail.com)}
\address[2]{Department of Mechatronics Engg., Manipal Institute of Technology, Karnataka, India - 576104 (email:diyaparekh0603@gmail.com)}
\address[3]{Department of Mechanical Engg., Manipal Institute of Technology, Karnataka, India - 576104 (email:thakur4pranav@gmail.com)}
\address[4]{Department of Data Science Engg., Manipal Institute of Technology, Karnataka, India - 576104 (email: yadavhrishikeshsingh@gmail.com)}
\address[5]{Department of Mechatronics Engg., Manipal Institute of Technology, Karnataka, India - 576104 (email: ramya.moorthy@manipal.edu)}
\address[6]{Department of Aeronautical \& Automobile Engg., Manipal Institute of Technology, Karnataka, India - 576104 (email:srinivas.g@manipal.edu)}

\tfootnote{*These authors share equal contribution\\ Project funded by Manipal Academy of Higher Education (MAHE) and undertaken by authors as members of thrustMIT Rocketry Team.}

\markboth
{Author \headeretal: Preparation of Papers for IEEE TRANSACTIONS and JOURNALS}
{Author \headeretal: Preparation of Papers for IEEE TRANSACTIONS and JOURNALS}

\corresp{Corresponding author: Srinivas G. (e-mail: srinivas.g@manipal.edu).}

\begin{abstract}
The use of parallel manipulators in aerospace engineering has gained significant attention due to their ability to provide improved stability and precision. This paper presents the design, control, and analysis of “ STEWIE ”, which is a three-degree-of-freedom (DoF) parallel manipulator robot developed by members of the thrustMIT rocketry team, as a payload stabilization mechanism for their sounding rocket, 'Altair'. The goal of the robot was to demonstrate the attitude control of the parallel plate against the continuous change in orientation experienced by the rocket during its flight, stabilizing the payloads. At the same time, the high gravitational forces (G-forces) and vibrations experienced by the sounding rocket are counteracted. A novel design of the mechanism, inspired by a standard Stewart platform, is proposed which was down-scaled to fit inside a 4U CubeSat within its space constraints. The robot uses three micro servo motors to actuate the links that control the alignment of the parallel plate. In addition to the actuation mechanism, a robust control system for its manipulation was developed for the robot. The robot represents a significant advancement in the field of space robotics in the aerospace industry by demonstrating the successful implementation of complex robotic mechanisms in small, confined spaces such as CubeSats, which are standard form factors for large payloads in the aerospace industry.
\end{abstract}

\begin{keywords}
Parallel Manipulators, Robotics, Payload, Sounding Rocket, Control Systems, Kinematics
\end{keywords}

\titlepgskip=-21pt

\maketitle

\section{Introduction}

\PARstart{T}{he} successful use of payloads in space missions depends largely on ensuring their stability and orientation control during the launch and spaceflight phases. Vibration motion, G-forces and dynamic disturbances caused by the rocket's orientation during flight path can affect the integrity and functionality of the payload. To address these challenges, parallel manipulators have emerged as a promising solution, offering the potential to counteract these external perturbations and maintain precise alignment.

The inspiration for the design of our robot STEWIE comes from the versatility and robustness of the six degrees-of-freedom (DoF) Stewart platform, which is widely used in various industrial and research applications. Our research aims to take advantage of the principles of the Stewart platform while simplifying the kinematic structure to create a three-degrees-of-freedom (DoF) parallel manipulator suitable for payload stabilization in thrustMIT's sounding rocket: Altair.

The main goal of this article is to present a comprehensive study of STEWIE's design, development and performance characteristics. We address the kinematic analysis of the manipulator and illustrate its ability to provide three degrees-of-freedom for precise positioning of the payload during rocket ascent and spaceflight. In addition, we are investigating the integration of different types of joints and micro-servo motors into STEWIE's structure to improve its parallel alignment with respect to the Earth's surface. We present the results of the various structural simulations carried out on STEWIE to ensure that it can function effectively in the harsh conditions it will experience during its flight inside the rocket.

We also propose a computationally in-expensive but robust controller which has high response speeds and high movement accuracy.

The research results show that STEWIE has exceptional potential as an effective payload stabilization system for sounding rockets such as Altair. The ability of the manipulator to actively counteract vibrational movements, G-forces and orientation changes ensures the integrity and reliable functionality of payloads during critical phases of space missions. Furthermore, its compact, lightweight and simple design contributes to the rocket's overall efficiency and improved payload capacity.

\subsection{Contributions of Research}
The main contributions of this paper to the field of robotics and aerospace are the following:-
\begin{itemize}
    \item A novel low-complexity, pwm-controlled actuator based payload stabilization mechanism for rocket flights.
    \item A computationally lightweight controller, demonstrating high accuracy and response time for critical robotic application especially in space industries.
    \item A simple and robust custom tuning method for pid-controllers.
    \item An open-source platform comprising mechanical design files (STEP format), details of electronic components, and real-time-operating-system (RTOS).
\end{itemize}

\subsection{Oranisation of Paper}
The paper is organized as follows: a comprehensive study on related work and recent development in the the field of robotics, especially for aerospace applications and in payloads has been presented in section II. The design of the robot detailing its geometry, kinematics involved, various components used, actuation mechanism, along with communication interfaces, sensors and circuitry design are described in section III. A detailed description of the control architecture including the controllers used, are explained in section IV. A comprehensive description of the tuning methods for the controllers is given in section V. Analysis of the simulations, and actual rocket flight results are presented in section VI. Finally, we present our conclusion in section VII.

\section{Related Work \& Recent Developments}
The utilization of Stewart platforms in aerospace applications has seen a surge in interest owing to their capability to provide precise control and stabilization in dynamic environments. Recent advancements in this field have seen various applications ranging from vibration control systems to robotic manipulators designed specifically for space missions. This section reviews significant advancements in the fields of robotics, Stewart platforms, and their application within the aerospace industry, focusing on the stabilization of payloads during rocket flights.

Stewart platforms, first introduced by Eric V. Gough (1947) [2] and D.Stewart et al. (1965) [1] are known for their high stiffness, force capabilities, and precise motion control and have been increasingly employed in aerospace applications to address challenges related to maintaining orientation and stability in dynamic environments. Research by Tsai et al. (1999) [3] and Melek et al. (2010) [4] have highlighted the mechanics and kinematics of Stewart platforms, emphasizing their suitability for tasks requiring high precision and complex motion control.

Wang et al. (2022) [7] explored the utilization of Stewart platforms in the low-frequency vibration characteristic testing of space truss deployable antennas mounted on satellites. Their study demonstrated the effectiveness of Stewart platforms in conducting vibration tests, crucial for assessing the structural integrity and performance of satellite components in space environments.

Preumont et al. (2007) [6] presented a pioneering work on a six-axis single-stage active vibration isolator based on a Stewart platform. This work introduced a novel approach to mitigating vibrations in aerospace systems, which is essential for improving the accuracy and reliability of sensitive instruments and payloads onboard spacecraft.

Liang et al. (2021) [5] contributed to the field with their research on the design and implementation of a high-precision Stewart platform tailored for a space camera. Their study focused on achieving precise positioning and stabilization of the camera platform, crucial for capturing high-quality images and conducting remote sensing operations in space.

In a related study, Liang et al. (2022) conducted kinematics and dynamics simulations of a Stewart platform, providing valuable insights into its motion characteristics and performance metrics. 

Furthermore, Glassner (2021) explored the concept of soft Stewart platforms tailored for robotic in-space assembly applications. This innovative approach introduces flexibility into the platform structure, enabling adaptability to varying task requirements and enhancing safety during assembly operations in space environments.

These works collectively highlight the diverse applications and recent advancements in Stewart platform technology within the aerospace industry. The versatility, precision, and adaptability of Stewart platforms make them indispensable for a wide range of aerospace missions, including satellite deployment, vibration control, payload stabilization, and in-space assembly operations.

\label{sec:related_work}

\section{Robot Design}
In this section we will discuss in detail the overall geometry of the mechanical structure, the forward kinematic calculations for the robot, the computing \& communication implemented, finally focusing on the leg design and the actuation mechanism of the robot.
\label{sec:robot-design}

\begin{figure}
    \centering
    \includegraphics[width=\linewidth]{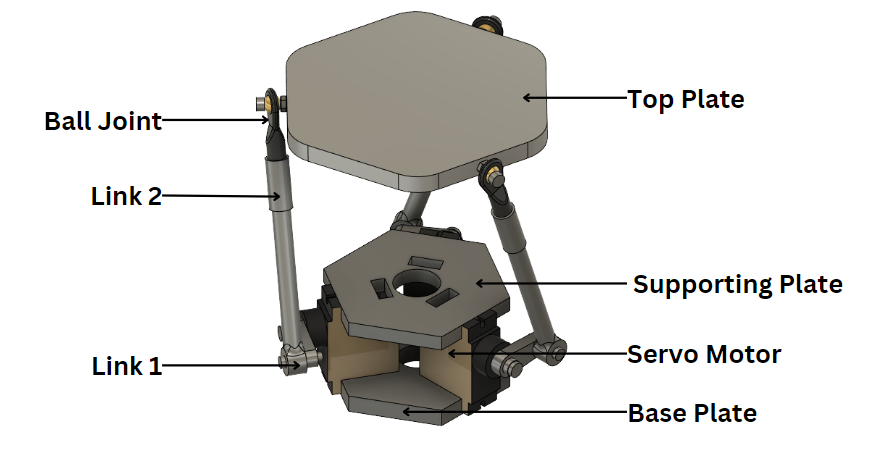}
    \caption{Computer Aided Design (CAD) of STEWIE Robot}
    \label{fig:STEWIE-cad}
\end{figure}

\subsection{Geometry of Robot}
The hardware design of STEWIE is inspired from a Stewart-Gough Platform [1]. The design philosophy behind the mechanism is based on compactness, light weight construction, ease of manufacturing, rapid repair and robust operation. 
\begin{table}[!htbp]
    \centering
    \caption{Physical Robot Parameters}
    \begin{tabular}{cccc}
    \hline
         Parameter&  Symbol & Value & Units\\
         \hline
         Mass & $m$ & 485 & g\\
         Minimum Body Height & $l_{body}$ & 110 & mm\\
         Maximum Body Height & $l_{body}$ & 125 & mm\\
         Body Width & $w_{body}$ & 85 & mm\\
         Leg Link Length & $l_1, l_2, l_3$& 79 & mm\\
         Maximum Load Capacity & - & 10.91 & kg-f\\
         \hline
    \end{tabular}
    
    \label{tab:robot_params}
\end{table}

The design of  the robot can be understood as an assembly of three sections, stationary base \& middle plates, 3 identical custom designed legs and a dynamic top plate where the payload to be stabilized sits (refer Fig. 1). STEWIE weighs around 485 $g$ with a maximum height of 125 $mm$, when fully extended and minimum height of 110 $m$, from ground to top plate depending on the positions of the legs. Each leg of the robot measures a total length of 79 $mm$ and has a maximum load capacity of 10.91 $kg-f$.

\subsubsection*{\underline{Degrees of Freedom}}
The degrees-of-freedom of STEWIE were calculated using Chebychev–Grübler–Kutzbach criterion:
\begin{equation}
\label{gubler-eq}
    F_{T} = 6(N-1-J) + \sum (F_{i} \cdot J_{i}) 
\end{equation}
Where, \\
$N$ : Number of links = 8 \\
$J$ : Number of joints = 9 \\
$F_{i}$ : Degrees of Freedom of a joint \\
$J_{i}$ : Number of joints with the same degree-of-freedom\\

Here, the summation term of (1) represents the effective constraints which are acting on the robot. In STEWIE, six revolute joints with one-degree-of-freedom and three ball joints with three-degrees-of-freedom each have been used. Hence, the summation term of (1) can be calculated as follows:
\begin{equation*}
\sum(Fi \cdot Ji) = (1 \cdot 6) + (3 \cdot 3) = 15
\end{equation*}

Therefore, on substituting the given values in the criterion given in (1), degrees-of-freedom calculated for STEWIE is calculated as $F_{T} = 3$.

\subsection{Computing and Communication}
\begin{figure}[!htbp]
    \centering
    \includegraphics[width=8cm]{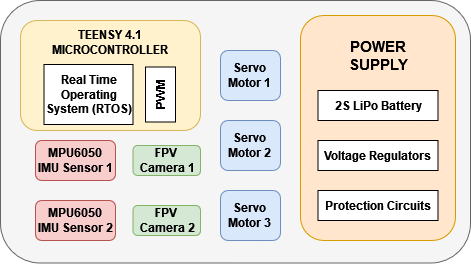}
    \caption{Electronics System Block Diagram}
    \label{fig:ele-sch}
\end{figure}
The robot was deployed inside a CubeSat, which was divided into two segments. The upper segment houses the robot along with two first-person-view (FPV) cameras to record the movement of the plates during the rocket flight from two different angles. For inertial measurement, a small printed-circuit-board (PCB) is mounted on the top plate of the robot which has a MPU6050 inertial-measurement-unit (IMU) on it for orientation data of the plate. A secondary IMU is mounted on the bottom plate of the robot, which remains stationary with respect to the frame of reference of the rocket during its flight. This secondary IMU helps to determine the orientation of the top plate in reference to the rocket's frame of reference. The bottom segment consists of majority of the electronics components, cable routing and two power distribution boards each for the two cameras. The power is supplied to the parallel manipulator robot through an extension from a 2S Lithium-Polymer (LiPo) battery with additional batteries for redundancy. The central processing unit of the robot is a Teensy 4.1 microcontroller, functioning on a FreeRTOS based real-time-operating-system (RTOS). The processor communicates with the IMU sensor daughter boards through high-speed Inter-Integrated-Circuit (I2C) interface with each IMU sensor on a different address. The servo motors are controlled by the microcontroller via pulse-width-modulated (PWM) signals generated by it (refer Fig. 2). This setup executes control strategies, including control of motors at frequencies as high as 200 Hz.

\subsection{Leg Design \& Actuation}
The robot comprises of three legs. Each leg of the robot is divided into three sections. The bottom section of the leg is a revolute joint which comprises a micro-servo motor placed at the base of each leg and a link-1 (refer Fig. 3) which connects the motors to the middle section of the leg. Link-1 is connected to the middle section of the leg through a nut-bolt mechanism, which acts as the second revolute joint. The middle section of the robot is a cylindrical rod which we call as Link-2 (refer Fig. 3). The top section of the leg comprises of a ball joint which connects the leg to the top plate of the robot. The motors are attached and sandwiched between the middle and the base plate of the robot to prevent them from moving during the actuation of the robot. The motor acts as the driver for the actuator and the joints above are driven by it, translating the motor's rotary motion into linear and angular displacement of the top plate. Each leg of the robot measures a total length of 79 $mm$ and weights around 26 $g$. 

\begin{figure}[!htbp]
    \centering
    \includegraphics[width=\linewidth]{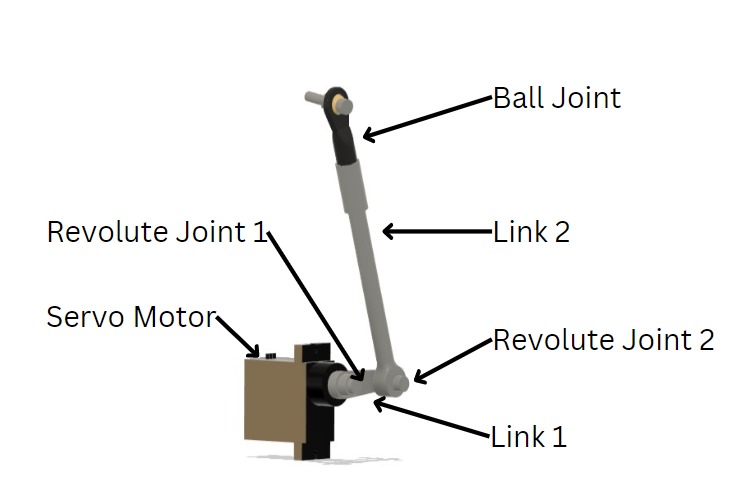}
    \caption{Leg Design of Robot}
    \label{fig:leg_design}
\end{figure}

During development of the robot, other potential methods of actuation were also explored such as pneumatic actuators, linear actuators in its early iterations. Though effective, but these early iterations were rejected due to unavailability of cost effective commercial actuators and space constraints inside the CubeSat.

\begin{table}[!htbp]
    \centering
    \caption{Specification of Components of Robot}
    \begin{tabular}{cccc}
    \hline
       Part  & Material & Weight & Units\\
         \hline
        Top Plate & ABS Plastic& 53 & g\\
        Middle Plate & Steel& 105 & g\\
        Base Plate & Steel & 85 & g\\
        Link 1 & Aluminium 6061 & 13 & g\\
        Link 2 & Aluminium 6061 & 4 & g\\
        Ball Joint & Plastic & 3 & g\\
         \hline
    \end{tabular}
    
    \label{tab:robot_specs}
\end{table}

\begin{table}[!htbp]
    \centering
    \caption{Micro Servo Motor Specifications}
    \begin{tabular}{ccc}
    \hline
       Parameter  & Value & Units\\
         \hline
        Model & SG-90 & \\
        Mass & 9.0 & g\\
        Motor Size & 22.2 & mm\\
        & 11.8 & mm\\
        Operating Speed & 0.1 & s/60 degree\\
        Operating Voltage & 4.8 & V\\
         Maximum Torque & 2.5 & Kg-cm\\
         \hline
    \end{tabular}
    
    \label{tab:motor_specs}
\end{table}

\section{Control Architecture}

\label{sec-control}
Our control architecture follows a combination of a custom robot state estimator which analyzes the orientation of the robot's top plate along the $\{X, Y, Z \}$ axes and performs forward kinematic calculations to determine the position matrix of the ball joints in the form of cartesian coordinates in the three-dimensional space, which is then passed as an input to three separate low-level controllers for each of the three legs (refer Fig. 6). The controllers generates the motor angles for each motor, by which the it should move to ensure that the top plate is parallel to the ground no matter the orientation of the robot or the rocket. This in-turn stabilizes the payload which is kept on the robot. 

This control approach allows for a decoupling of the control process, simplifying the design and implementation.

\begin{figure}
    \centering
    \includegraphics[width=\linewidth]{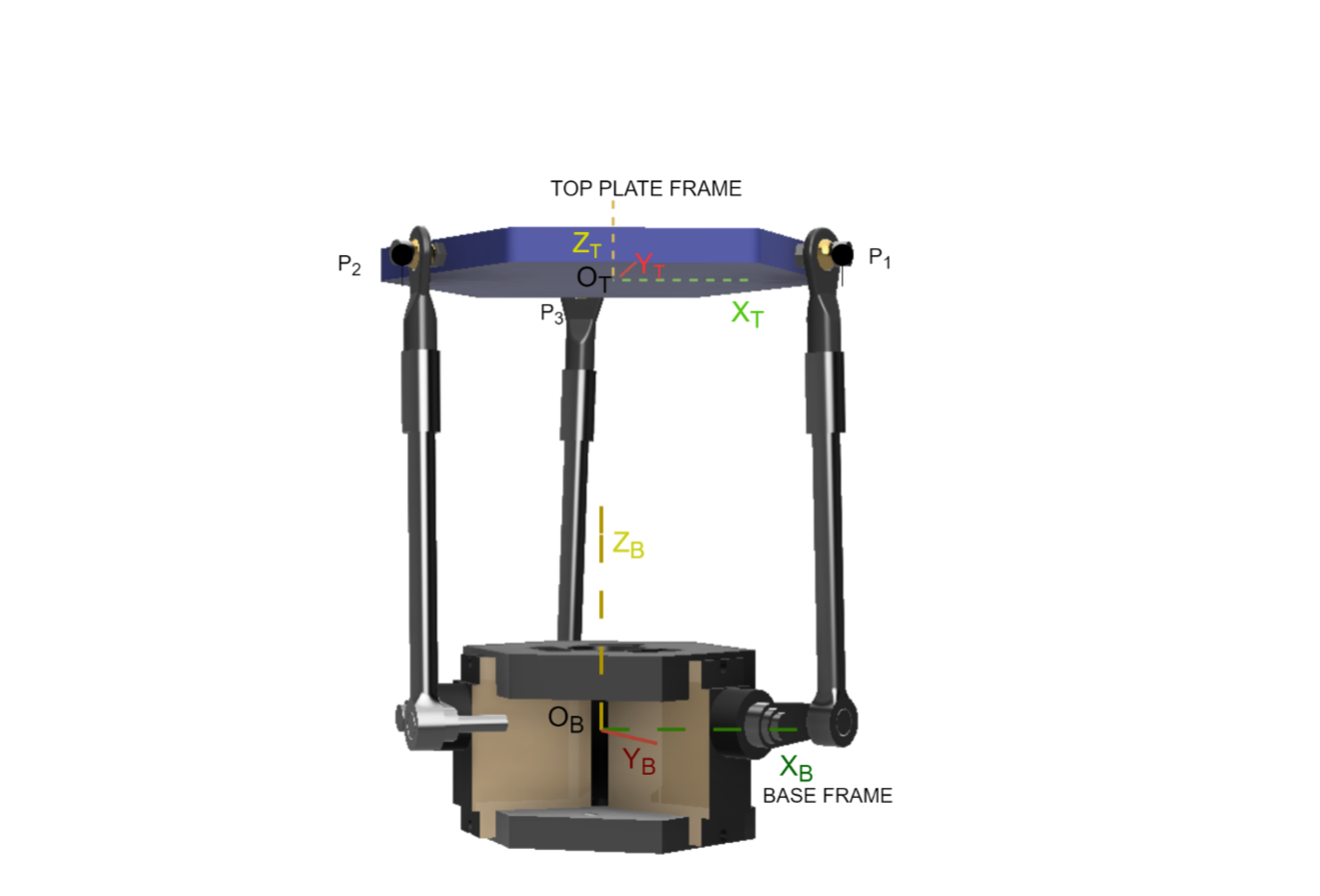}
    \caption{Frame of References in Robot}
    \label{fig:frames}
\end{figure}

\subsection{Robot State Estimator}
\textbf{Robot's State}:  The state of the robot at any given time, is a three-dimensional vector $S_t = \{X_p, Y_p, Z_p \}$ consisting of the position coordinates of the ball joints with respect to the the base frame (refer Fig. 4) along the X, Y \& Z axis at which the legs are connected to the plate through the ball joints.

Forward Kinematic calculations were done on the robot to determine the final position matrix of the ball joints. The IMU sensor which is attached to the top plate of the robot has a three-axis gyroscope and can give orientation data of the top plate in the form of angle deviations from three axes as known as Yaw, Pitch and Roll. This information was utilized to calculate the position of ball joints in the base frame. Here, the 'base frame' is defined as the initial frame located at the centre of the three motors (refer Fig. 4). The base frame is at a fixed distance from the frame of reference of the top plate throughout the flight irrespective of the position of the motors. 

Position matrix of top plate with-respect-to base plate frame of the robot was presented as  ${}^{B}T_{T}$, where,  $\{ \theta_{X}, \theta_{Y}, \theta_{Z} \}$ are the angle deviations of the top plate along from the $X, Y, Z $ axes obtained from the Yaw-Pitch-Roll data from the IMU and ${}^BZ_T$ is the fixed distance between the top plate frame and base frame. The position matrix of ball joint with respect to top plate frame was represented as ${}^{T}T_{P}$, where, $\{x, y, z \}$ are the coordinates of the ball joints in the three-dimensional space in the frame of reference of the top plate. 

To find position matrix of ball joint with respect to base frame, ${}^{B}T_{T}$ and ${}^{T}T_{P}$ were multiplied.
\begin{equation}
    {}^{B}T_{P} = {}^{T}T_{P} \cdot {}^{B}T_{T}
\end{equation}
Refer to Fig. 5 for detailed matrix multiplications.

\begin{figure*}
    \begin{align*}
        \mathbf{{}^{T}T_{P}} =\begin{pmatrix} 
    1 & 0 & 0 & x \\
    0 & 1 & 0 & y \\
    0 & 0 & 1 & z \\
    0 & 0 & 0 & 1 \\
\end{pmatrix}
    \end{align*}
\end{figure*}

\begin{figure*}[]
\begin{align*}
\mathbf{{}^BT_T} &=\begin{pmatrix} 
    \cos\theta_Z \cdot \cos\theta_Y & -\sin\theta_Z \cdot \cos\theta_X + \cos\theta_Z \cdot \sin\theta_Y \cdot \sin\theta_X & \sin\theta_Z \cdot \sin\theta_X + \cos\theta_Z \cdot \sin\theta_Y \cdot \cos\theta_X &  0\\
    \sin\theta_Z \cdot \cos\theta_Y & \cos\theta_Z \cdot \cos\theta_X + \sin\theta_Z \cdot \sin\theta_Y \cdot \sin\theta_X & -\cos\theta_Z \cdot \sin\theta_X + \sin\theta_Z \cdot \sin\theta_Y \cdot \cos\theta_X & 0 \\
    -\sin\theta_Y & \cos\theta_Y \cdot \sin\theta_X &  \cos\theta_Y \cdot \cos\theta_X & {}^BZ_T \\
    0 & 0 & 0 & 1 \\
\end{pmatrix}
\end{align*}
\end{figure*}

\begin{figure*}[]
\begin{align*}
\mathbf{{}^BT_P} &=\begin{pmatrix} 
    \cos\theta_Z \cdot \cos\theta_Y & -\sin\theta_Z \cdot \cos\theta_X + \cos\theta_Z \cdot \sin\theta_Y \cdot \sin\theta_X & \sin\theta_Z \cdot \sin\theta_X + \cos\theta_Z \cdot \sin\theta_Y \cdot \cos\theta_X &  X_{p} \\
    \sin\theta_Z \cdot \cos\theta_Y & \cos\theta_Z \cdot \cos\theta_X + \sin\theta_Z \cdot \sin\theta_Y \cdot \sin\theta_X & -\cos\theta_Z \cdot \sin\theta_X + \sin\theta_Z \cdot \sin\theta_Y \cdot \cos\theta_X & Y_{p} \\
    -\sin\theta_Y & \cos\theta_Y \cdot \sin\theta_X &  \cos\theta_Y \cdot \cos\theta_X & Z_{p} \\
    0 & 0 & 0 & 1 \\
\end{pmatrix}
\end{align*}
\caption{Position Matrices of Robot}
\label{pos-mat}
\end{figure*}

\begin{figure*}
    \begin{align*}
Where,\\
X_{p} &= (\cos\theta_Z \cdot \cos\theta_Y) \cdot x + (-\sin\theta_Z \cdot \cos\theta_X + \cos\theta_Z \cdot \sin\theta_Y \cdot \sin\theta_X) \cdot y + (\sin\theta_Z \cdot \sin\theta_X + \cos\theta_Z \cdot \sin\theta_Y \cdot \cos\theta_X) \cdot z\\
Y_{p} &= (\sin\theta_Z \cdot \cos\theta_Y) \cdot x + (\cos\theta_Z \cdot \cos\theta_X + \sin\theta_Z \cdot \sin\theta_Y \cdot \sin\theta_X) \cdot y + (-\cos\theta_Z \cdot \sin\theta_X + \sin\theta_Z \cdot \sin\theta_Y \cdot \cos\theta_X) \cdot z\\
Z_{p} &= (-\sin\theta_Y) \cdot x + (\cos\theta_Y \cdot \sin\theta_X) \cdot y + (\cos\theta_Y \cdot \cos\theta_X) \cdot z + Z_T^B
    \end{align*}
\end{figure*}

\subsection{Low Level Controller}
Approaching the robot's locomotion challenge from a control theory perspective, we aim to develop a policy capable of parameterizing the relationship between the angular positions of the servo motors and the orientation of robot's top plate in the three-dimensional space furthermore analyzing how changing the angular position of the servo changes the orientation of the top plate. These parameterized value act as inputs for the controllers, which, in turn, incorporates a regulatory mechanism based on the orientation state of each leg.

\begin{figure*}[!htbp]
\centering
    \includegraphics[width=\linewidth]{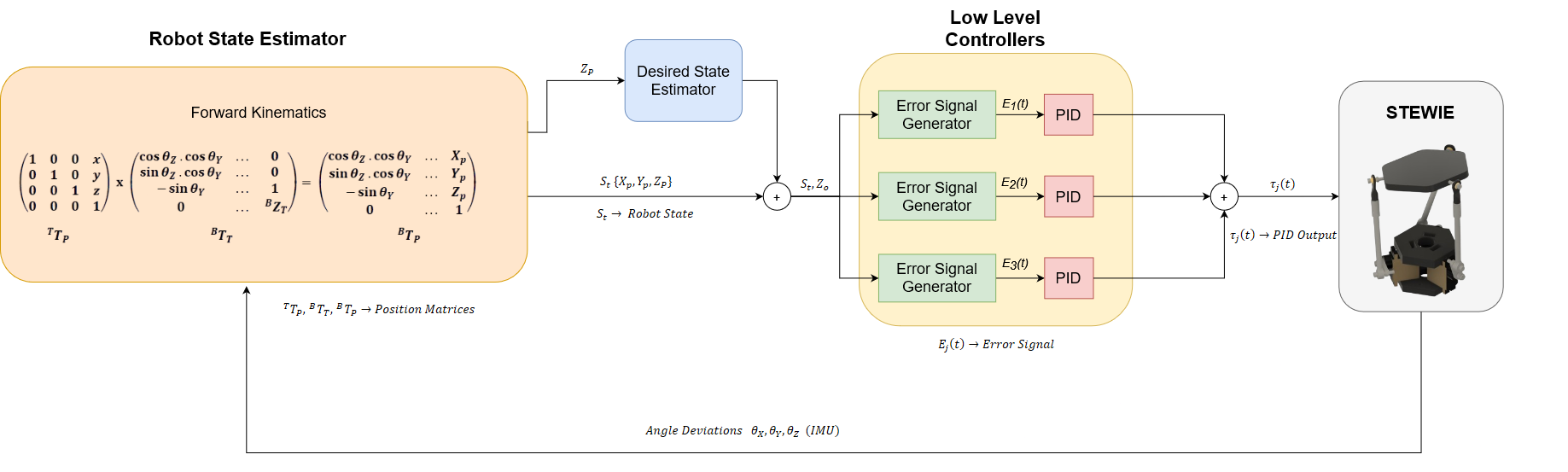}
    \caption{Control Architecture}
    \label{fig:cotrl_archi}
\end{figure*}

\textbf{Desired State}: The desired state of the robot is one in which the top plate should be parallel to the ground. In our case, when the top plate is parallel to the ground, it is in absolute horizontal orientation with respect to the ground surface and at that instance, the $z-$coordinate of all the ball joints are same. Therefore, the controller will be regulating the motors such that their $z-$coordinate of all the motors are same to achieve horizontal orientation of the top plate with respect to ground. 

\textbf{Error Signal}: At any given state, the errors signal ($E_j$) for the three controllers are calculated by taking the absolute difference between between the desired z-coordinate, $Z_o$, also known as set-point (SP) and the actual z-coordinates of the three points of the ball joints, $Z_{p_1}, Z_{p_2}, Z_{p_3}$, also known as process-variables (PV) in control theory.

\begin{equation}
    E_j = |Z_o - Z_{p_j}| \ \ \forall \ \ j = 1, 2, 3
\end{equation}

The desired z-coordinate ($Z_o$) for the controller is chosen as the median of the actual z-coordinates of the three ball joints. 

\begin{equation}
    Z_o = \ median\{Z_{p_1}, Z_{p_2}, Z_{p_3} \}
\end{equation}

This also reduces the computational complexity of the model as, at given state, controller calculations for only two of three motors have to be done because the leg having the median z-coordinate will remain stationary.

The motor level controllers tracks the z-coordinates of the plate points at every time instant '$t$' using a Proportional-Integral-Derivative (PID) controller with appropriate gains $K_p$, $K_i$, and $K_d$ for each of the three controllers as follows:

\begin{multline}
\tau_{j}(t) = K_{p_j} \cdot E_j(t) + K_{i_j}\cdot \sum^{t-1}_{k = 0} E_j(k) + K_d \cdot \dot E_j(t)\\ 
    \ \ \ \forall \ \ j = 1, 2, 3
\end{multline}Where,
\begin{itemize}
    \item[] $\tau_{t_j}$ : PID output of the $j^{th}$ controller at time instant $t$.
    \item[] $E_{t_j}$ : Error of the $j^{th}$ controller at time instant $t$. Ref. (3)
    \item[] $K_{p_j}$ : Proportional gain of the $j^{th}$ controller.
    \item[] $K_{i_j}$ : Integral gain of the $j^{th}$ controller.
    \item[] $K_{d_j}$ : Derivative gain of the $j^{th}$ controller.
\end{itemize}

\section{Tuning of Controllers}
The effective tuning of PID controllers is crucial for achieving optimal performance in robotic system, particularly those with complex dynamics such as parallel manipulators. In this section, we will discuss the methods employed for tuning the PID controllers. We utilized three different methods: a custom tuning method and two heuristic tuning methods, Ziegler-Nicholas, proposed by J.G. Ziegler \& N.B. Nicholas (1942) [13] and the Cohen-Coon Method, proposed by G.H. Cohen and G.A. Coon (1953) [14]. 

In all the methods, to get system response signal, the process-variable which are the z-coordinates $Z_{p_1}, Z_{p_2}, Z_{p_3}$ for the three ball joints and set-point, which is the desired z-coordinate $Z_o$;  were plotted against time $t$. Final values of PID gains were chosen based on the comparison results of the different methods.  

\subsection{Custom Tuning Method}
This method was an modified version of the trial-error method of tuning a controller. The method involved performing open-loop test on the controller to determine the process characteristics of the controller followed by manually adjusting the gains depending on the results of the open-loop test. We used a random value of the set-point during each iteration as the reference signal.

The process involved following steps:
\subsubsection{Perform Open-loop Test}
The controllers were configured in manual mode and a step-change was introduced as the output signal. The process-variable (PV) curve response was monitored continuously and it was determine the that controller gave a self-regulating characteristic with pure $1^{st}$ order lag. Hence, it was inferred that the proportional gain was dominant for the controller had to be aggressive along with a nominal integral gain. The derivative gain could be used or avoided depending on the results in the future.

\subsubsection{Manually Adjusting Gains}
After recognising the characteristics of the controller, the gains were initialized with values of $K_p$ \& $K_i = 0.5$ and $K_d = 0$. Post initialization, they are manually adjusted based on rules given in Table 4 to achieve the best compromise between stability and quick response of the system.


\subsection{Ziegler-Nichols Method (Closed Loop)}
Ziegler-Nicholas closed-loop is a heuristic method for tuning controllers in which certain parameters of the system such as Ultimate Gain ($K_u$) and Critical Period ($\tau_u$) are determined manually through experimentation. Following which, the $K_p, K_i, K_d$ gains can be calculated through pre-determined rules based on the ultimate gain and the critical period given in Table 5.

\begin{table}[H]
    \centering
    \caption{Ziegler-Nicholas Method Tuning Rule}
    \begin{tabular}{|c|c|c|c|c|c|}
    \hline
        \backslashbox{\textbf{Type}}{\textbf{Gain}} & $\textbf{K_p}$ & $\textbf{T_i}$ & $\textbf{K_i}$ & $\textbf{T_d}$ & $\textbf{K_d}$\\
         \hline
        P & $K_u / 2$ & - & - & - & -\\
         \hline
        PI & $K_u / 2.2$ & $\tau_u / 1.2$ & $K_p / T_i$ & - & - \\
         \hline
        PID & $K_u / 1.7$ & $\tau_u / 2$ & $K_p / T_i$ & $\tau_u / 8$ & $K_p * T_d$\\
         \hline
    \end{tabular}
    \label{tab:zn-parameter}
\end{table}

The process involved the following steps:
\begin{description}
    \item[Step 1]: Keeping the integral gain ($K_i$) and derivative gain ($K_d$) as zero, proportional gain ($K_p$) is increased until the system starts to oscillate continuously at a constant amplitude. The gain at this point is called the Ultimate Gain ($K_u$). The calculated values of ultimate gains for the three controllers is given in Table 6.

    \item[Step 2]: Critical Period ($\tau_u$) is determined by measuring the period of oscillation at the ultimate gain. The calculated values of critical periods for the three controllers is given in Table 6.
    \item[Step 3]: Using a set of pre-defined formulas presented by Ziegler-Nicholas given in Table 5, the values of $K_p, K_i, K_d$ gain were calculated. Final values of the gains are given in Table 6.
\end{description}

\begin{table}[H]
    \centering
    \caption{Gains calculated through Ziegler-Nicholas Method}
    \begin{tabular}{|c|c|c|c|c|c|}
    \hline
        \textbf{Type of Controller} & $\textbf{K_u}$ & $\textbf{\tau_u}$ & $\textbf{K_p}$ & $\textbf{T_i}$ & $\textbf{T_d}$\\
         \hline
        Leg 1 PID Controller & 6.63 & 0.24 & 3.98 & 0.12 &0.03\\
         \hline
        Leg 2 PID Controller & 4.65 & 0.24 & 2.79 & 0.10 &0.03\\
         \hline
        Leg 3 PID Controller & 7.92 & 0.16 & 4.75 & 0.08 &0.02\\
         \hline
    \end{tabular}
    \label{tab:zn-parameter}
\end{table}

\subsection{Cohen-Coon Method}
Cohen-Coon method is an extension of the Ziegler-Nicholas Method, but uses more information from the system in the process. The system's response is modelled to a step change as a first-order response plus dead time using this method. From the response, three parameters: $K, \tau_m$ and $\tau_d$ are founded. $K$ is the output steady state divided by the input step change, $\tau_m$ is the effective time constant of the first-order response, and $\tau_d$ is the dead time.

\begin{equation}
    \tau_m = \frac{3}{2} (t_2 - t_1)
\end{equation}

\begin{equation}
    \tau_d = t_2 - \tau_m
\end{equation}
To verify the results, according the Cohen Coon rules, the dead time should be less than two times the length of the time constant [1]. The Cohen Coon relations for calculating the PID gains is given in Table 7.
\renewcommand{\arraystretch}{2.5}
\begin{table}[H]
    \centering
    \caption{Cohen-Coon Method Tuning Relations}
    \begin{tabular}{|m{5em}|c|c|c|}
    \hline
        \textbf{Type of Controller} & $\textbf{K_p}$ & $\textbf{T_i}$ & $\textbf{T_d}$\\
         \hline
        P & $\frac{\tau_m}{K\tau_d}(1+\frac{\tau_d}{3\tau_m})$ & - & -\\
         \hline
        PI & $\frac{\tau_m}{K\tau_d}(0.9+\frac{\tau_d}{12\tau_m})$ & $\tau_d(\frac{30+3\tau_d/\tau_m}{9+20\tau_d/\tau_m})$ & -\\
         \hline
         PD & $\frac{\tau_m}{K\tau_d}(1.25+\frac{\tau_d}{6\tau_m})$ & - & $\tau_d(\frac{6-2\tau_d/\tau_m}{22+3\tau_d/\tau_m})$\\
         \hline
        PID & $\frac{\tau_m}{K\tau_d}(1+\frac{\tau_d}{3\tau_m})$ & $\tau_d(\frac{32+6\tau_d/\tau_m}{13+8\tau_d/\tau_m})$ & $\tau_d(\frac{4}{11+2\tau_d/\tau_m})$\\
         \hline
    \end{tabular}
    \label{tab:zn-parameter}
\end{table}

The obtained system parameters and calculated systems gains are given in Table 8.

\begin{table}[H]
    \centering
    \caption{System parameters calculated through Cohen-Coon Method}
    \begin{tabular}{|c|c|c|c|c|c|c|c|c|}
    \hline
        \textbf{Controller} & $\textbf{K}$ & $\textbf{\tau_d}$ & $\textbf{\tau_m}$ & $\textbf{K_p}$ & $\textbf{T_i}$  &$\textbf{T_d}$\\
         \hline
        Leg 1 PID Controller & 2.9 & 0.07 & 0.47 & 2.85 &0.43 & 0.08\\
         \hline
        Leg 2 PID Controller & 3.92 & 0.03 & 0.37 & 2.15 &0.25& 0.06\\
         \hline
        Leg 3 PID Controller & 1.71 & 0.05 & 0.55 & 4.58 &0.55& 0.05\\
         \hline
    \end{tabular}
    \label{tab:zn-parameter}
\end{table}

The system response curves for tuning PID controllers using the above given methods is given in Fig. 7. 

\begin{figure*}[!htbp]
    \centering
    \includegraphics[width=\linewidth]{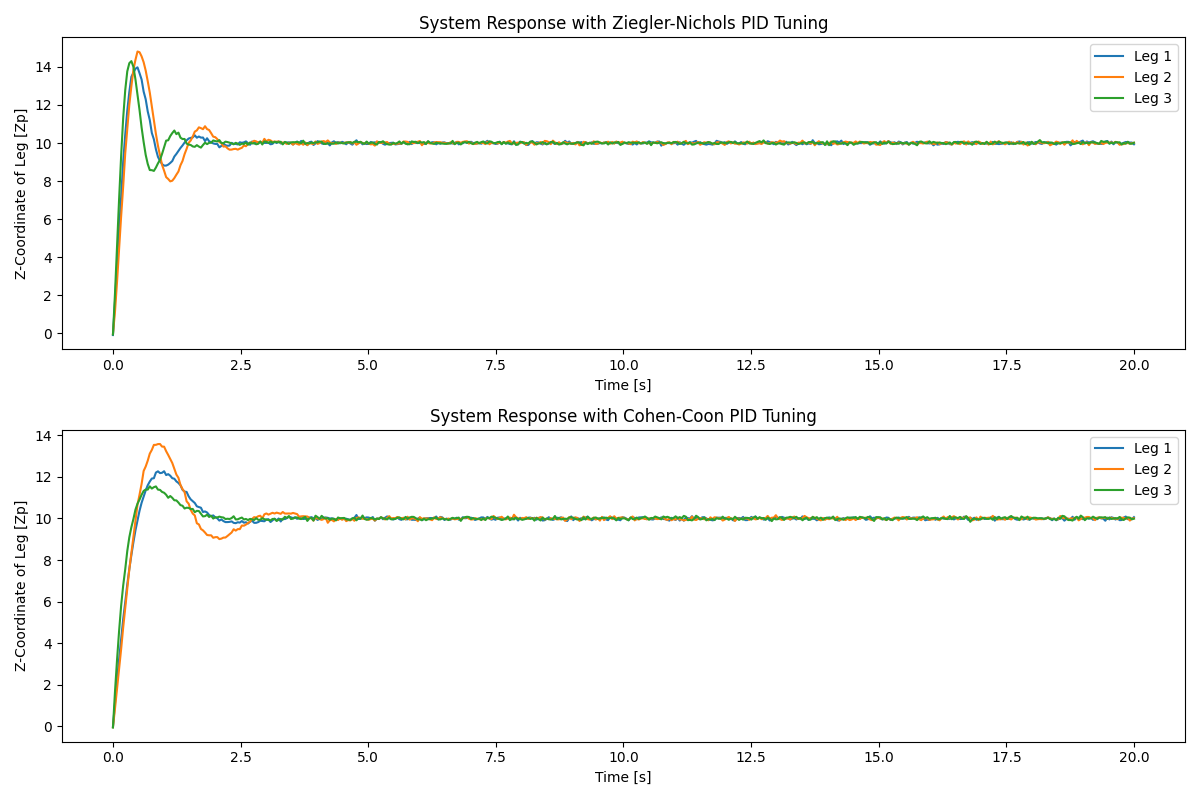}
    \caption{System Response Curves for PID Controllers with desired Zo = 10}
    \label{fig:enter-label}
\end{figure*}

\section{Results \& Discussions}
\label{sec-results}

\subsection{Simulations}
\subsubsection{Modal Analysis}
Necessary modal analysis simulation was performed on the robot to observe its behaviour under high vibrational forces. The formations and stress zones were identified at different natural frequencies and accordingly the design of the robot was also modified to provide more support in high stress areas. Based on the analysis, the robot was finally designed to sustain a maximum of 16 G-forces under continuous orientation change during the rocket flight.

The simulations were performed for five mode shape results for a total deformation. As expected, the maximum stress was found to be at the joints in the link-2 and the ball joint as they were externally joined by a transition fit but sufficient length was not present to hold on. Stress regions were also observed at the sharp edges of the top plate and the link-2, which weren't filleted to reduce stress concentration. However, the edges were filled and filleted later to least possible diameter to justify the stress concentration. The maximum total deformation observed was 261.57 $mm$ at a frequency of 761.13 $Hz$. Since, the natural frequency of the rocket during its flight was between 70 - 100 $Hz$, we were in a safe zone with a decent amount of factor of safety.

\subsubsection{Static Structural Simulation}
It was necessary to perform static structural simulations on the robot to observe the effects of downward acting load on the robot, experienced during the rocket flight. To carry out the simulations, the 3D CAD model was imported into the Ansys software as a STEP file. Appropriate materials were assigned to the respective parts of the robot and the necessary regions were given the property of contact regions bonded with no separation. To define the threads region, local coordinate system was assigned to the center of the bolts. 

Further, meshing was performed to the model, which was a combination of Tet and Hex mesh elements. The regions where the load effects were anticipated to be higher like the links, Hex mesh was assigned to such regions for accurate results and the regions where the loads were anticipated to be lesser, Tet mesh was assigned to reduce overall complexity and compute time of the simulation. There were a total of 0.25 $million$ elements in quadratic order with a size of 0.75 $mm$.  

The force applied to the simulation was calculated as following:-
The maximum torque which can be delivered by the servo motor was 10$g/cm$ or 1$kg/m$. The length of each link was 15 $mm$, hence through simple torque calculation, the maximum force on each link was 66.66 $N$ in the ideal condition. Considering losses and practical efficiency of 85\% for the motors, the maximum force on each link can be calculated as 56$N$ or 5.6$Kg$. Adding a factor of safety of 1.5 for maximum load, each link could lift a maximum weight of 3.8$Kgs$, therefore making it 11.4$Kgs$ for the whole robot which converts to approximately, 107$N$. 

During the rocket's flight, the maximum downward G-forces acting on the rocket was 12G's which is equivalent to 117$N$. Majority of this downward force is experienced by the frame and the structure of the rocket body namely the stringers and the bulkheads. The payload is also protected by its own supporting plates and the walls of the CubeSat. Thus, the net effect of these downward forces acting on the payload is significantly reduced, approximately up to 50\% of its original value. 

Therefore, final value of the G-force acting on the payload is considered as 59$N$. The weight of the top plate was itself 5$N$ and the weight of the payload which was kept on the top plate to be balanced was 6$N$. Adding all these forces gives a total of 70$N$ of acting in the downward direction, which is well below the maximum weight that the links can handle.

For simulation purposes, the maximum value of down force was considered and applied at the centre of the top plate, the fixed support at the bolt holes and the base plates. Displacement support to the ball joints were also added. Simulations were done for total deformation and equivalent stress to locate the high stress regions. The maximum deformations of 4.098e-4 $mm$ were observed at the edges of the top plate, where there were no bolt holes. The maximum equivalent stress of 6.637 $MPa$ was observed at the points where the ball joints were attached to the top plates. 

Observing these results, we concluded that there was no major concern regarding the design as every parameters was within the expected range even with aspect ratio being only 7. The simulation proved the design's load bearing capacity.

\begin{figure}[]
    \centering

       \includegraphics[width=\linewidth]{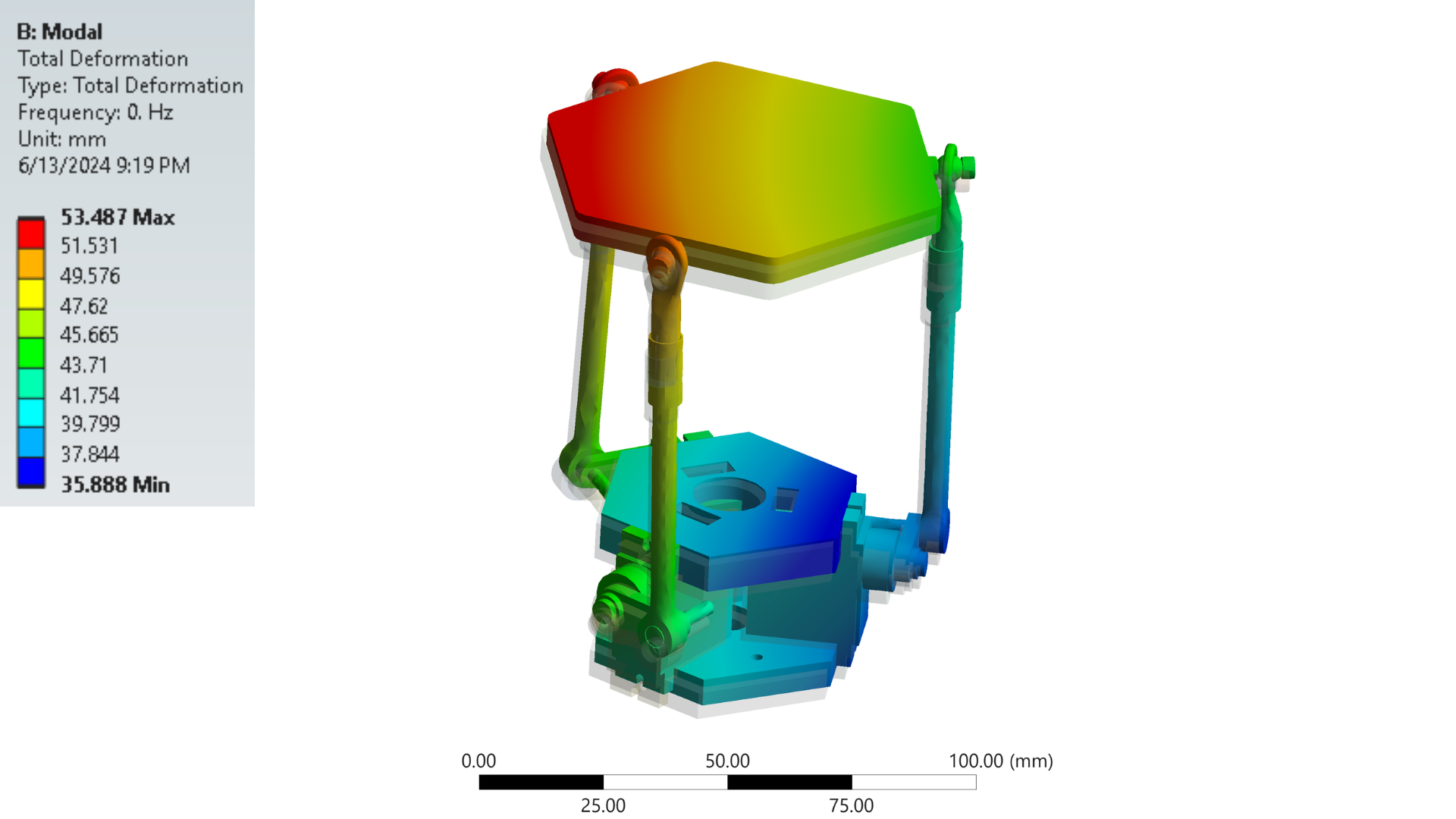}
    \caption{Modal Analysis Total Deformation 1}
    \label{fig:enter-label} 

    \includegraphics[width=\linewidth]{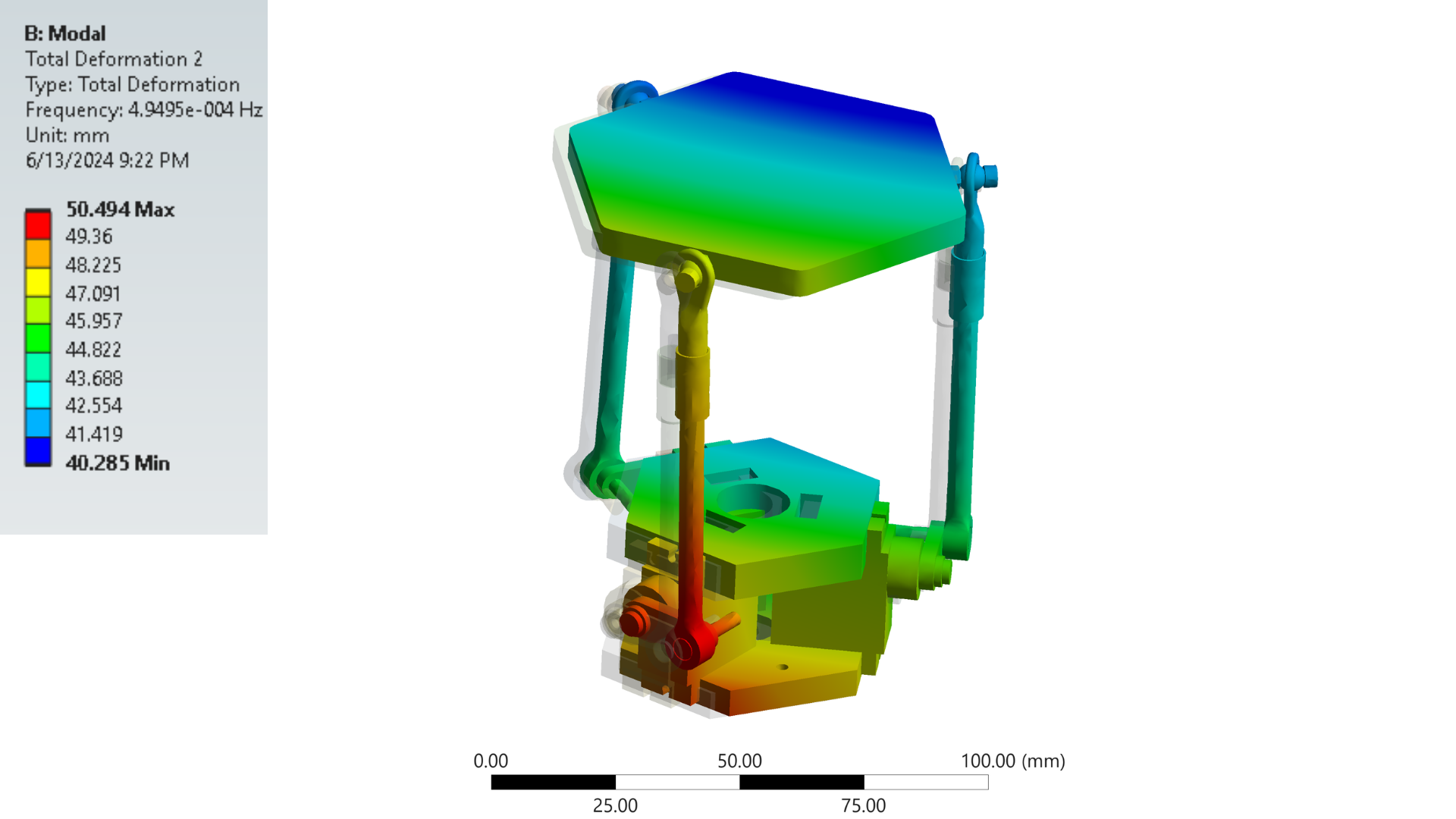}
    \caption{Modal Analysis Total Deformation 2}
    \label{fig:enter-label}
    
\end{figure}

\begin{figure}[!htbp]
    \centering
    \includegraphics[width=\linewidth]{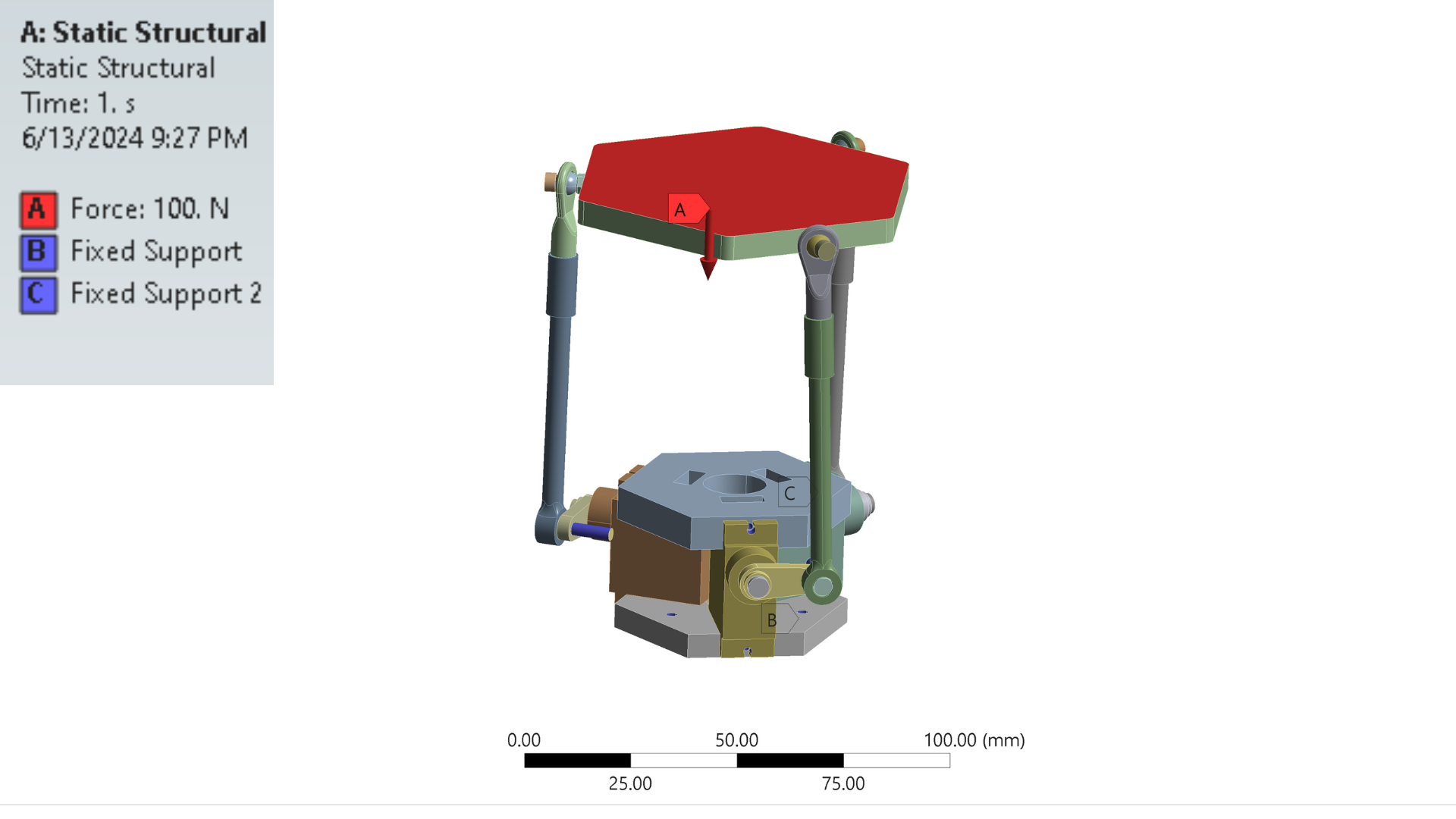}
    \caption{Static Structural Parameters}
    \label{fig:enter-label}

\end{figure}
\begin{figure}[!htbp]
    \includegraphics[width=\linewidth]{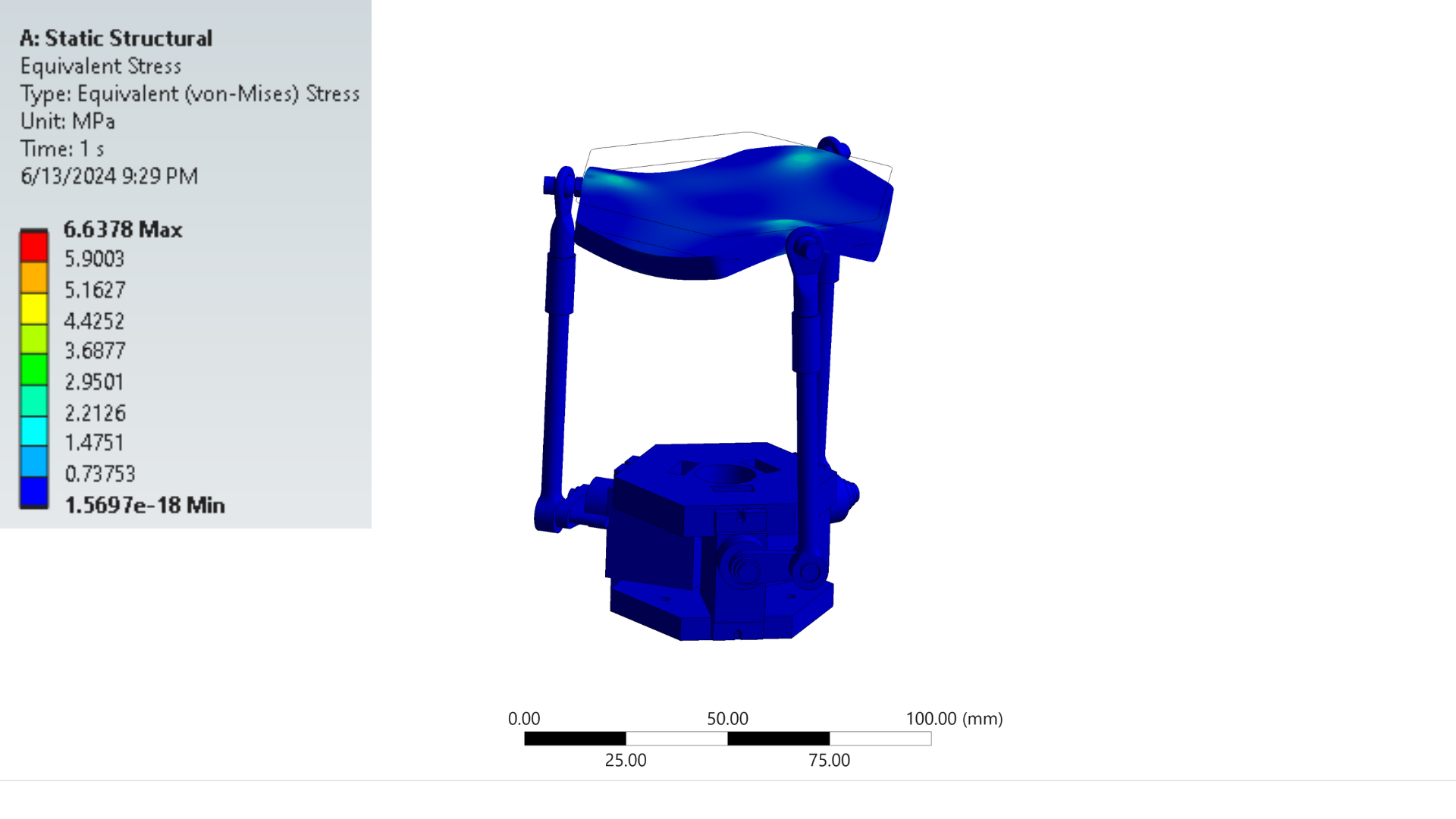}
    \caption{Static Structural Equivalent Stress}
    \label{fig:enter-label}
\end{figure}
\begin{figure}[!htbp]
    \includegraphics[width=\linewidth]{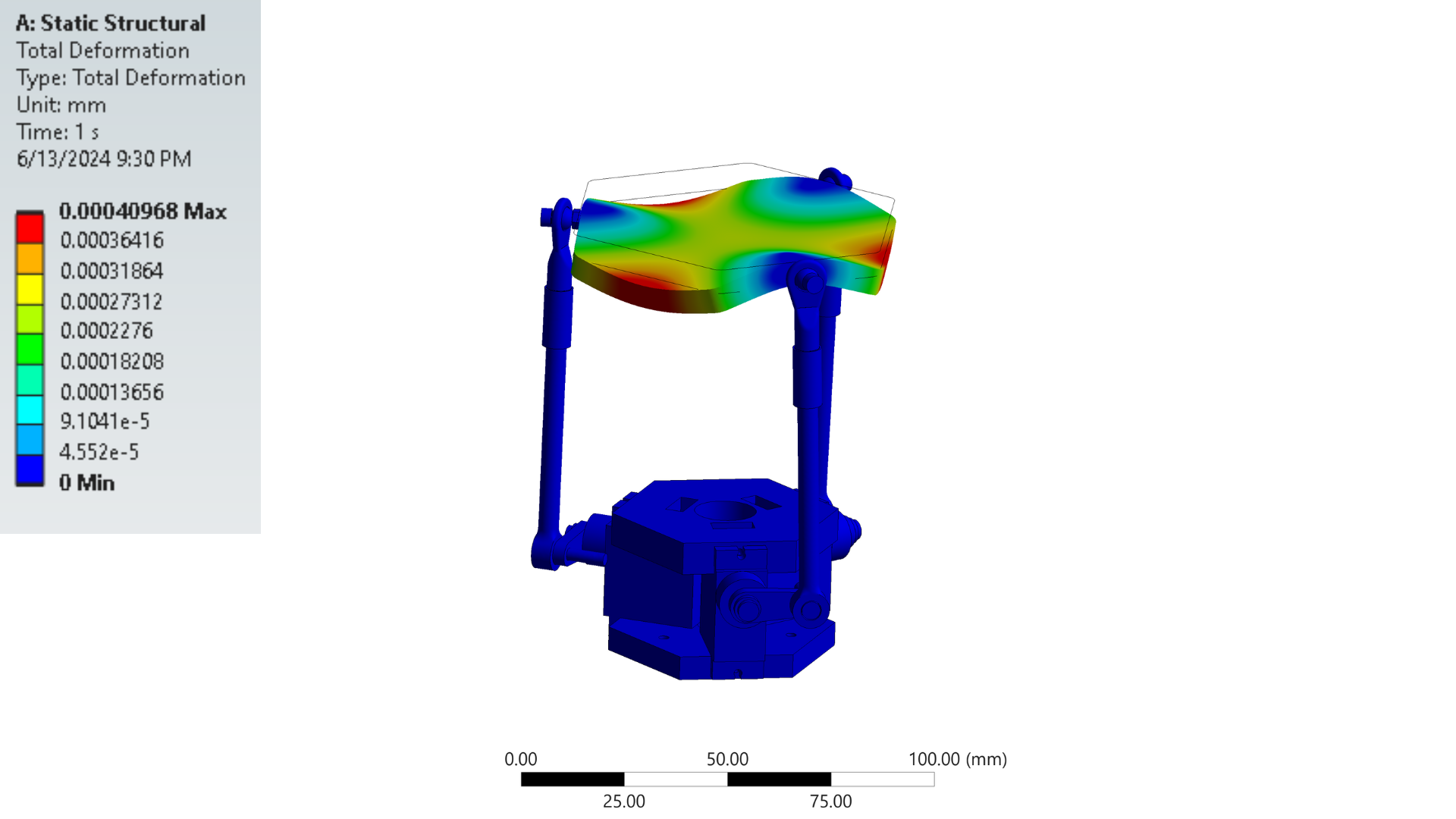}
    \caption{Static Structural Total Deformation}
    \label{fig:enter-label}
\end{figure}

\subsection{Rocket Flight Results}
The robot was launched on an actual sounding rocket for testing. Throughout the operation of the robot, the IMU data was being logged onto an external SD Card by the microcontroller. Post rocket flight, the data was accessed, processed and cleaned for analysis. The Yaw-Pitch-Roll data obtained from the SD Card is presented in Fig. 8. The data concludes that the top plate of the robot remained almost horizontal with respect to the ground throughout the rocket flight and the angle deviations from the absolute zero degrees were minimum.

\begin{figure*}[b]
    \centering
    \includegraphics[width=\linewidth]{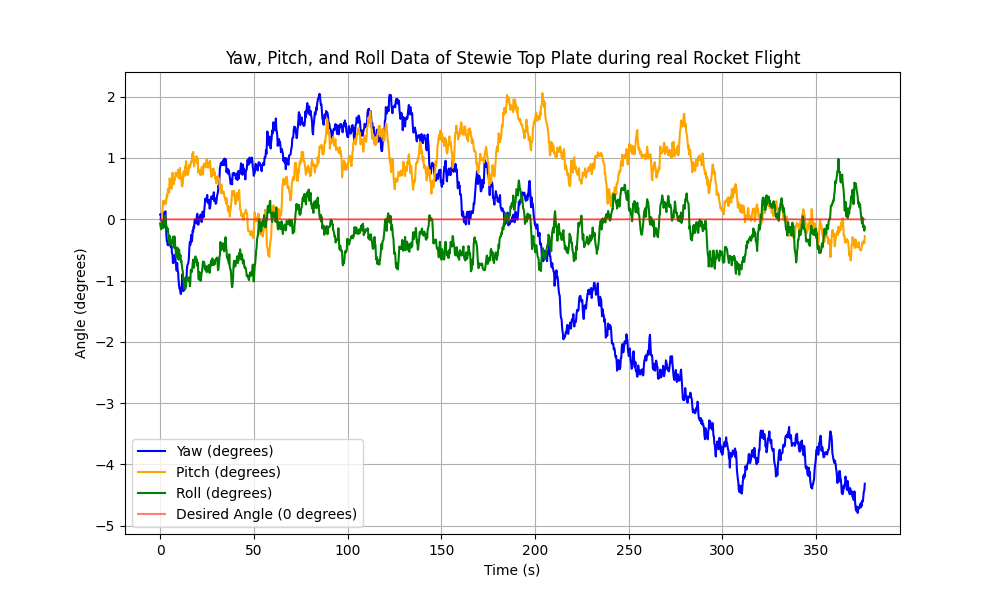}
    \caption{Yaw - Pitch - Roll data gathered from IMU sensor}
    \label{fig:IMU-Data}
\end{figure*}

\section{Conclusions}
\label{sec-conlcusion}
The development and implementation of the STEWIE payload stabilization mechanism represent a significant advancement in the field of aerospace engineering. This research has demonstrated the feasibility and effectiveness of using a three-degree-of-freedom (DoF) parallel manipulator for maintaining the stability and orientation of payloads during rocket flights. The design, inspired by the Stewart platform, successfully addressed the challenges posed by high G-forces, vibrations, and dynamic disturbances experienced during rocket ascents.

Extensive simulations and analyses were conducted to evaluate the robot's performance and robustness in dynamic environments. Modal analysis and static structural simulations confirmed the design's ability to withstand up to 16-G forces during continuous orientation changes. Rocket flight tests validated the effectiveness of the control system, with the top plate maintaining near-horizontal alignment throughout the flight.

Our results indicate that STEWIE's novel low-complexity, PWM-controlled actuator and lightweight control system provide exceptional precision and responsiveness, crucial for space applications. The robust performance of STEWIE in various simulations and actual flight tests confirms its potential to enhance payload stability, ensuring the integrity and functionality of critical instruments during missions.

The compact and lightweight design of STEWIE also contributes to the overall efficiency of the rocket, allowing for improved payload capacity without compromising performance. This research paves the way for future innovations in payload stabilization mechanisms, suggesting that similar systems can be adapted for a variety of aerospace applications, including CubeSats and other small-scale payloads.

Future research directions include exploring the use of advanced materials, optimizing the kinematic structure, and investigating the integration of additional sensors for enhanced control and monitoring capabilities.

In conclusion, STEWIE exemplifies the integration of advanced robotics and control systems in aerospace engineering, providing a reliable solution for payload stabilization in dynamic and harsh environments. Future work will focus on further refining the design, exploring the use of alternative materials and actuation methods, and expanding the application of this technology to other types of space missions.

\section{Acknowledgement}
We sincerely thank thrustMIT, Manipal Institute of Technology, and the Manipal Academy of Higher Education for their
invaluable support and resources that greatly facilitated the successful completion of this research.

\begin{IEEEbiography}[{\includegraphics[width=1.15in,height=1.25in,clip,keepaspectratio]{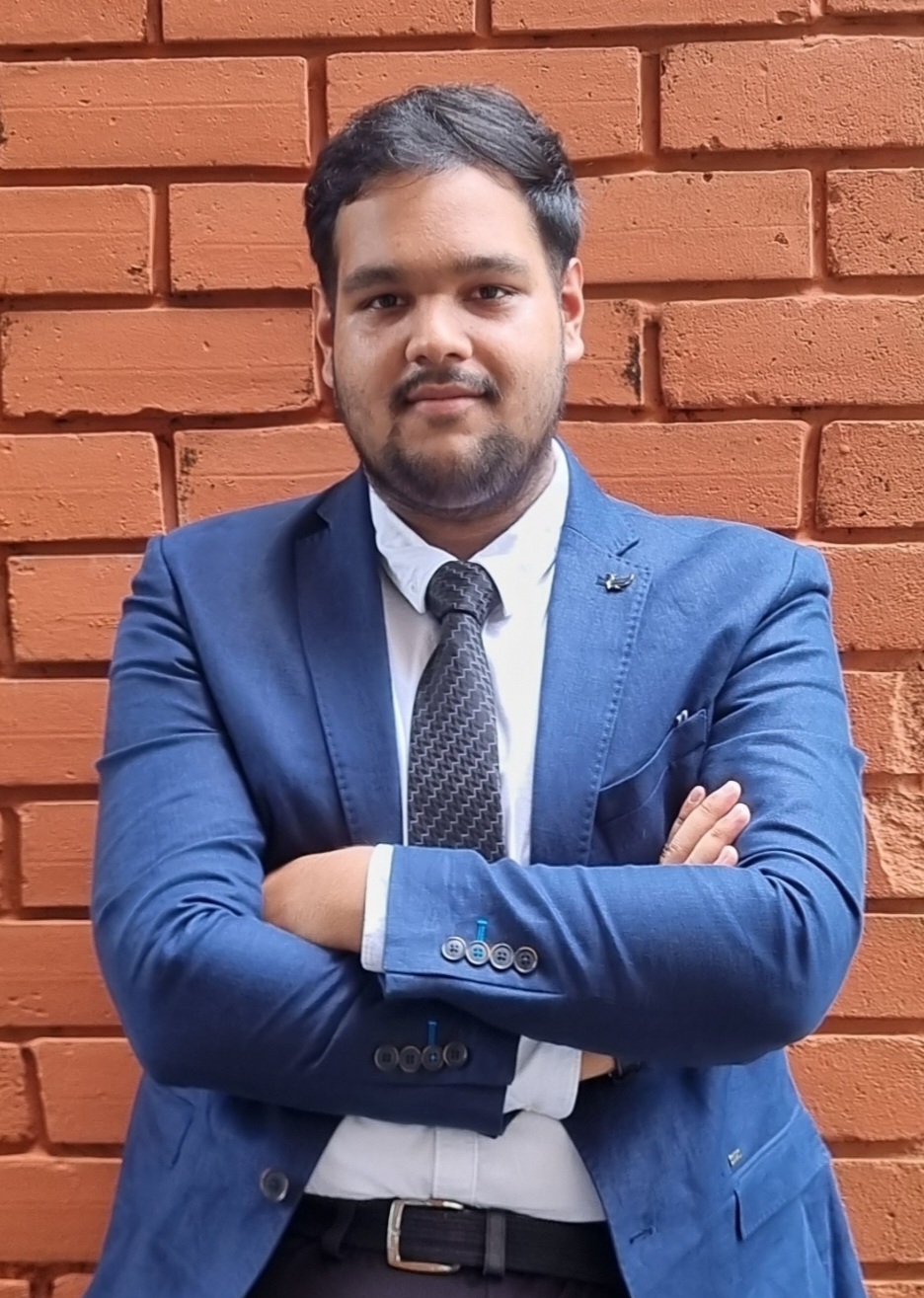}}]{Utkarsh Anand} is a senior undergraduate pursuing his B.Tech in Electrical and Electronics Engineering from Manipal Institute of Technology, Karnataka, India, graduating in 2024. 

From 2021 to 2023, he was member of the thrustMIT student rocketry team as Head of the Payload division and Launch Operations Lead for the team. Apart from payloads, he has also worked on developing novel deployment mechanism for payloads in sounding rockets as well as Deep Learning based control mechanisms for optimized control of airbrakes in the rocket. Between 2022 and 2023, he has also worked as a Research Interns at several other prestigious institutions such as Indian Institute of Sciences (IISc), Indian Institute of Technology Roorkee (IITR) and Defence Research \& Development Organization (DRDO). His research interests include legged robotics, controller design for robotic applications in space and reinforcement learning.

Mr. Utkarsh Anand was a recipient of the prestigious IASc‐INSA‐NASI Summer Research Fellowship by Indian Academy of Sciences in 2022. He was a member of Association of Computing Machinery (ACM) from 2022 to 2023.
\end{IEEEbiography}

\begin{IEEEbiography}[{\includegraphics[width=1.05in,height=1.25in,clip,keepaspectratio]{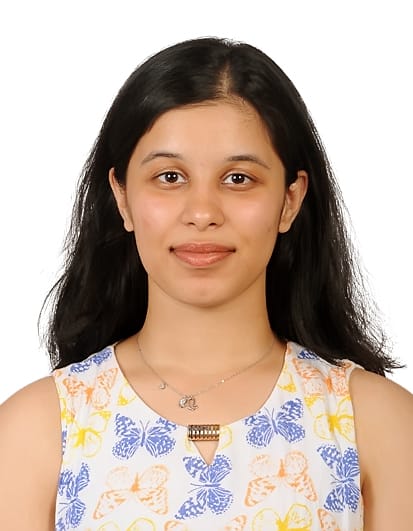}}]{Diya Parekh} is a senior undergraduate pursuing her B.Tech in Mechatronics Engineering from Manipal Institute of Technology, Karnataka, India, graduating in 2024. 

From 2021 to 2023, she was a member of Payload division at thustMIT student rocketry team, a constituent of Manipal Academy of Higher Education. She has also worked on other projects like Direct Air Capture(DAC) in payload. Apart from developing payloads, she has also worked several other projects such as Automated drilling machine, pick and place pneumatic arm, automatic dam shutter control system, soft gripper. Her research interests include design, kinematics and control of robots
\end{IEEEbiography}

\begin{IEEEbiography}[{\includegraphics[width=1.05in,height=1.25in,clip,keepaspectratio]{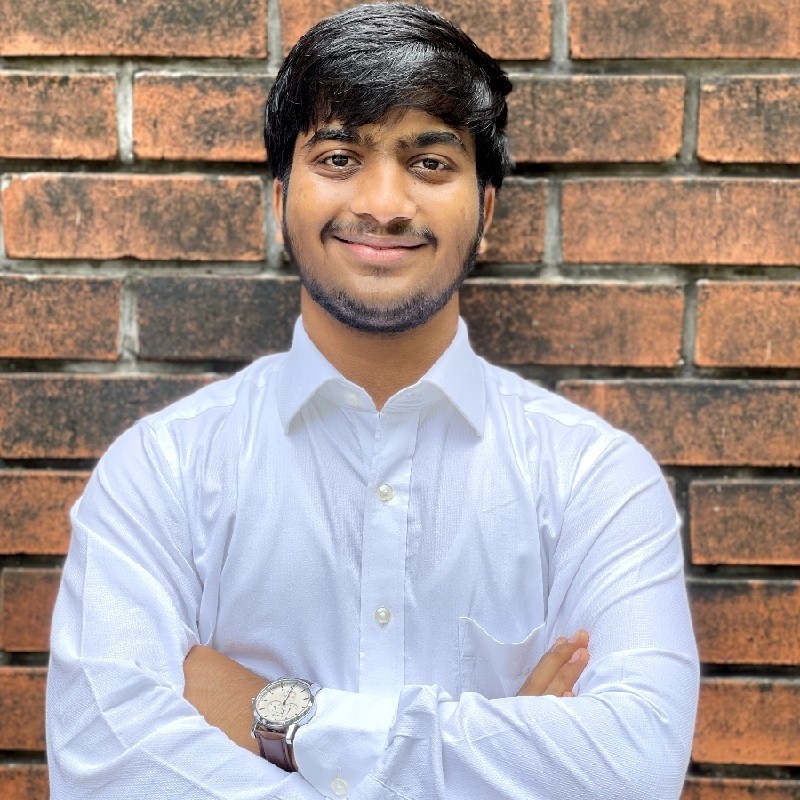}}]{Thakur Pranav G. Singh} is senior undergraduate, pursuing B.Tech in Mechanical Engineering from Manipal Institute of Technology, Karnataka, India, graduating in 2024.

From 2021 to 2023, he was a senior member of Payload and Structures division at the thrustMIT student rocketry team, a constituent of Manipal Academy of Higher Education. In 2023, he worked as an intern at the prestigious Defence Research and Development Laboratory (DRDL) a laboratory of Defence Research and Development Organization (DRDO), working on projects to develop a micro thrust stand to support micro-thrusters and to develop a pressure regulator for the same. He has earlier worked on projects like spectroscopy, Carbon Nano Tubes(CNT), Direct Air Capture (DAC) in payload and designing of a Radial deployment mechanism in sounding rockets.
\end{IEEEbiography}

\begin{IEEEbiography}[{\includegraphics[width=1.05in,height=1.25in,clip,keepaspectratio]{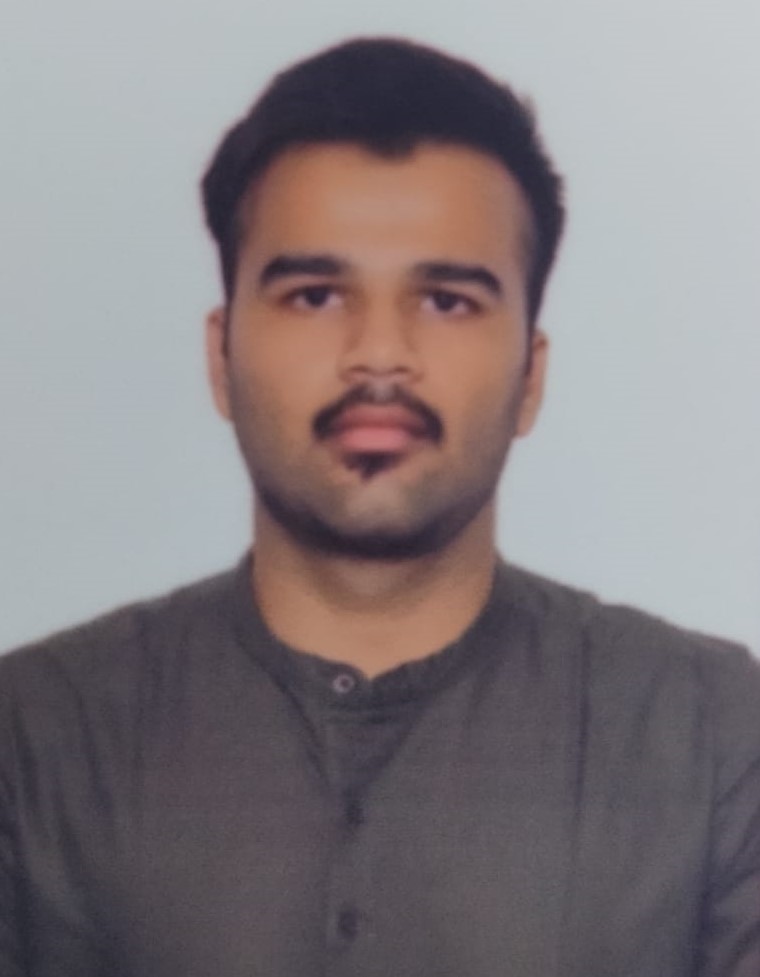}}]{Hrishikesh Singh Yadav} is an undergraduate pursuing his B.Tech in Data Science and Engineering from Manipal Institute of Technology, Manipal. He has been a member of the thrustMIT student rocketry team since 2022 and is currently serving as the Head of the Payload division.

He has been invited to present papers at the 6th GeoAI at the 31st ACM SIGSPATIAL International Conference on Advances in Geographic Information Systems 2023, held in Hamburg, Germany, and at the 2023 Asia Conference on Computer Vision, Image Processing, and Pattern Recognition (CVIPPR) held in Phuket, Thailand.

Hrishikesh is currently working as a Project Technical Lead at Standard Trading Company (STC) and has served as Project Director at Eon Space Labs. Additionally, he has worked as an SDE intern at the Lal Bahadur Shastri National Academy of Administration (LBSNAA), NIC.

He received a scholarship to attend the ACM SIGSPATIAL conference and is a current member of the Association of Computing Machinery (ACM).
\end{IEEEbiography}

\begin{IEEEbiography}[{\includegraphics[width=1.05in,height=1.25in,clip,keepaspectratio]{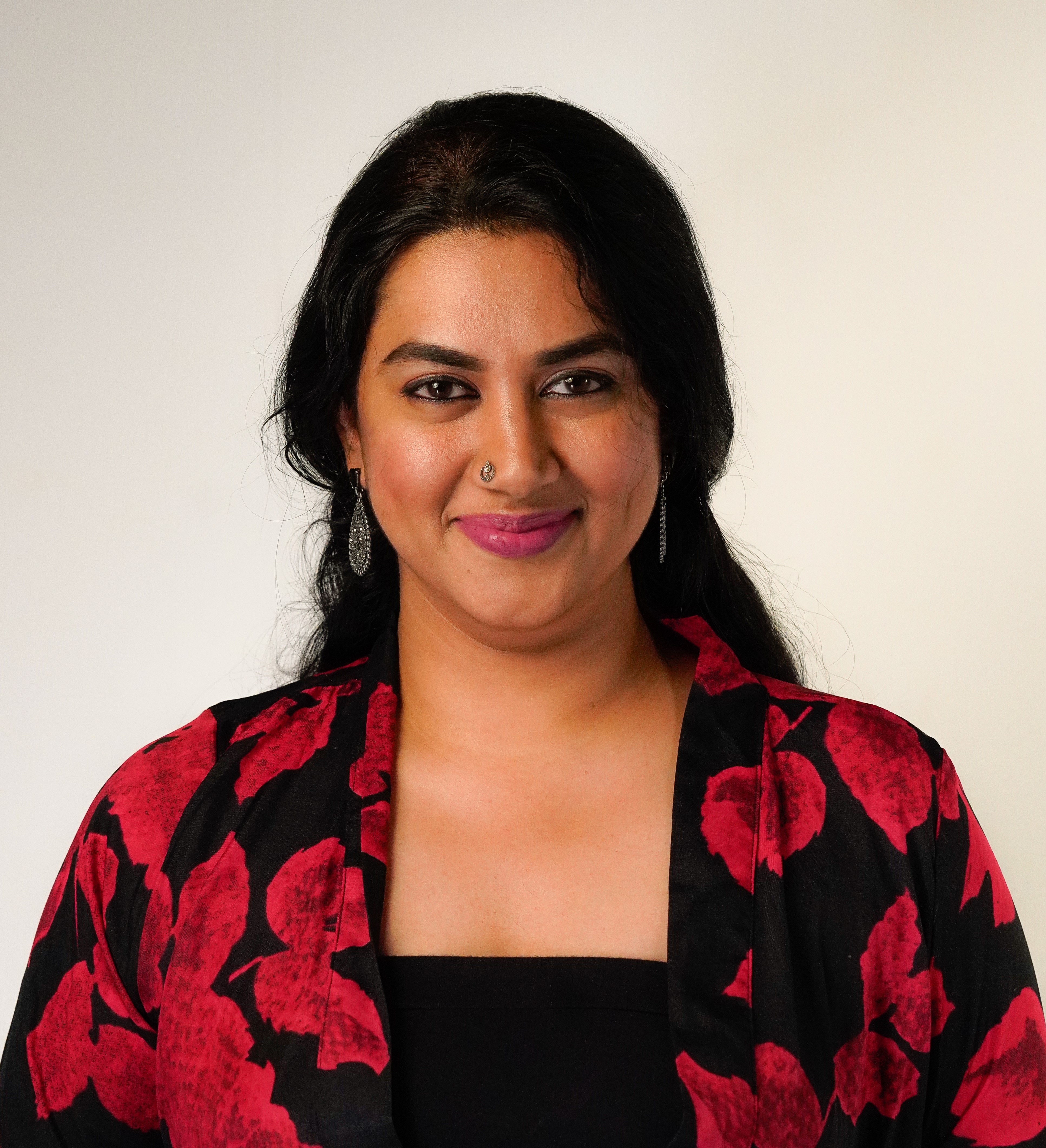}}]{Dr. Ramya S Moorthy} is the Founder Director of Nimaya Robotics Pvt. Ltd. from Chennai and has spent the last eight years of her life working towards a singular goal- to help children with special needs. 

Nimaya Robotics is a one-of-its-kind pioneering company that employs the use of robotics in the training of psychomotor \& cognitive skills for Autism Spectrum Disorder and other multiple disabilities. 

Dr. Ramya developed Suprayoga, an I-o-T based cloud monitored devices and training methodology designed to augment existing occupational therapy programs that helps accelerate the rate of learning by more than 60\% in children with ASD.

For Dr. Ramya the rationale for starting a company that works with children with Autism Spectrum Disorder was very simple- if her device could help even one child out there live a life that is not limited by their disorder, then she would do everything in her power to help that child. 

Dr. Ramya’s journey has been a whirlwind of events and emotions, with her living through a fire accident, losing her voice briefly, and the fire wrecking her life’s work. And yet, she has  shown phenomenal strength by getting back up and crossing the finish line, because she knew she had a purpose in life after that one phone call. It was from a special educator of a child with whom she had worked with in the past delightfully informing her that a child who was initially diagnosed with almost nil psychomotor skills at the start of the training, was able to remember and carry out every skill he had learnt even after a year. This planted the seed for Nimaya in Dr Ramya's heart. 

Since that day, there has been no looking back!

Dr. Ramya is an INSPIRE Fellow, IIS Fellow 2018, University Gold Medalist – M.Tech Robotics. She has Bachelor’s in Electrical and Electronics Engineering. Her work has also been published in several International Scientific Journals on Robotics. Few of her other experiences in Robotics are in the field of industrial robotics. She has been a guest speaker in several national and international panels/conferences and universities. Dr. Ramya S Moorthy is a recipient of the Devi Awards 2023, by Indian Express, Women Transforming India Award 2021, by NITI Aayog, amongst many other accolades.

Nimaya Robotics has also been awarded “Innovation of the year” Award by The Association of Lady Entrepreneurs (ALEAP), Telangana.

Dr. Ramya S Moorthy is also currently teaching at Manipal Institute of Technology, Manipal as an Asst. Prof.

\end{IEEEbiography}

\EOD

\end{document}